\documentclass[review]{elsarticle}

\usepackage{hyperref}
\journal{arXiv}

\usepackage[utf8]{inputenc} %
\usepackage[T1]{fontenc}    %
\usepackage{hyperref}       %
\usepackage{url}            %
\usepackage{booktabs}       %
\usepackage{amsfonts}       %
\usepackage{nicefrac}       %
\usepackage{microtype}      %
\usepackage{xcolor}         %

\usepackage{amssymb}
\usepackage{stmaryrd}
\usepackage{xfrac}
\usepackage{float}
\usepackage{multirow}
\usepackage{amsmath}
\usepackage{subcaption}
\usepackage{bm}
\usepackage{enumitem}
\usepackage{multicol}
\usepackage[
  ruled,
  vlined,
  linesnumbered,
]{algorithm2e}

\SetCommentSty{algcommentfont}
\usepackage{cleveref}
\crefformat{footnote}{#2\footnotemark[#1]#3}
\newcommand{\ra}[1]{\renewcommand{\arraystretch}{#1}}

\newcommand{\explainerStyle}[1]{{\small\texttt{#1}}}
\newcommand{\SHAP}{\explainerStyle{SHAP}}
\newcommand{\SHAPR}{\explainerStyle{SHAPR}}
\newcommand{\LIME}{\explainerStyle{LIME}}
\newcommand{\PDP}{\explainerStyle{PDP}}
\newcommand{\MAPLE}{\explainerStyle{MAPLE}}

\begin{document}

\begin{frontmatter}

\title{A Framework for Evaluating \textit{Post Hoc} Feature-Additive Explainers}
\author{Zachariah Carmichael\corref{mycorrespondingauthor}}
\cortext[mycorrespondingauthor]{Corresponding author}
\ead{zcarmich@nd.edu}
\author{Walter J. Scheirer}
\ead{walter.scheirer@nd.edu}
\address{Department of Computer Science and Engineering, University of Notre Dame\\384 Fitzpatrick Hall, Notre Dame, Indiana, 46556, United States of America}

\begin{abstract}
Many applications of data-driven models demand transparency of decisions, especially in health care, criminal justice, and other high-stakes environments. Modern trends in machine learning research have led to algorithms that are increasingly intricate to the degree that they are considered to be black boxes. In an effort to reduce the opacity of decisions, methods have been proposed to construe the inner workings of such models in a human-comprehensible manner. These \emph{post hoc} techniques are described as being universal explainers -- capable of faithfully augmenting decisions with algorithmic insight. Unfortunately, there is little agreement about what constitutes a ``good'' explanation. Moreover, current methods of explanation evaluation are derived from either subjective or proxy means. In this work, we propose a framework for the evaluation of \textit{post hoc} explainers on ground truth that is directly derived from the additive structure of a model. We demonstrate the efficacy of the framework in understanding explainers by evaluating popular explainers, \textit{e.g.}, \LIME{} and \SHAP{}, on thousands of synthetic and several real-world tasks.
The framework unveils that explanations may be accurate but misattribute the importance of individual features.
\end{abstract}


\end{frontmatter}

\section{Introduction}\label{sec:introduction}

The black box nature of many AI systems
obscures decision-making processes. This conceals biases, dubious correlations, and other deficiencies, which come to light especially in the cases of dataset shift~\cite{quionero-candelaDatasetShiftMachine2009}
and adversarial examples~\cite{DBLP:journals/corr/SzegedyZSBEGF13,hendrycksNaturalAdversarialExamples2019}.
Furthermore, this disparity in transparency has led to unexpected prejudices in real-world applications, including health care and criminal justice~\cite{oneilWeaponsMathDestruction2016,buolamwiniGenderShadesIntersectional2018}.
If this is not motivation enough for reconciliation,
current and forthcoming regulations will require transparency and a \textit{right to explanation}\footnote{It should be noted that current regulations, including the EU GDPR, have been argued to not actually guarantee such a right~\cite{10.1093/idpl/ipx005}.}, such as the EU General Data Protection Regulation~(GDPR)~\cite{EU-GDPR},
the US draft
``Guidance for Regulation of [AI] Applications''~\cite{us2020regulation},
and a United Nations prospective publication~\cite{unitednationsHogenhout2021}.

In endeavors to instill trust in these systems, an abundance of explainable AI (XAI) techniques have been introduced. However, these proposed techniques not only approach similar but distinct facets of interpretability -- they also associate notions of interpretability (and explanation) with different meanings. Several prevalent taxonomies have aimed to disentangle some of the perplexity of the language and concepts
\cite{doshi-velezConsiderationsEvaluationGeneralization2018,DBLP:conf/dsaa/GilpinBYBSK18,liptonMythos2018}.
Nonetheless, there is little consensus as to what interpretability is and, consequentially, how it may be evaluated. Existing efforts to assess the caliber of an XAI method rely on subjective or proxy means, which demonstrably mislead
or emphasize plausibility over fidelity~\cite{bucincaProxyTasksSubjective2020,DBLP:conf/dsaa/GilpinBYBSK18,herman2017}.
Understanding the fidelity of XAI methods is especially important as they
are
applied to various high stakes domains within both research and industry,
such as medical diagnosis, credit risk analysis, and epidemiological modeling~\cite{bhattExplainableMachineLearning2020,elshawiInterpretabilityMachineLearningbased2019,shapCOVID}.

In this study, we address the absence of truly objective evaluation
for \textit{post hoc} explanation methods. Specifically, our evaluation addresses the explanation aspect of XAI system evaluation as opposed to the aspects related to the user mental model~\cite{lofstrom2022meta}; if an explainer proves unfaithful within our framework, then it should not be user-facing regardless of whether users trust it.
Our work presents the following contributions:
\begin{itemize}[noitemsep,nolistsep]%
    \item We construct a test bed for the evaluation of feature-additive \textit{post hoc} explanations against objective \textit{ground truth} and establish the framework as a powerful tool for the comprehensive analysis of explainers.
    \item To facilitate evaluation, we propose an algorithm, \textsc{MatchEffects}, that directly maps a broad class of models to \textit{post hoc} explanations.
    \item We evaluate the popular \textit{post hoc} explainers \LIME{}~\cite{ribeiroWhyShouldTrust2016}, \SHAP{}~\cite{lundbergUnifiedApproachInterpreting2017}, \SHAPR{}~\cite{aasExplainingIndividualPredictions2020}, \PDP{}~\cite{Friedman2001}, and \MAPLE{}~\cite{plumbModelAgnosticSupervised2018}
    on thousands of synthetic tasks and models, as well as with multiple types of learned models on several prominent real-world datasets.
    \item We demonstrate that although \SHAP{}
    outperforms the other explainers, all explainers begin to fail in the presence of higher-dimensional data and on models with higher-order interactions and more interaction effects.
    \item We showcase interesting trends, \textit{e.g.}, that \textit{plausible} explanations are not necessarily \textit{faithful} and the precarity of poor explanations.
\end{itemize}

\section{Background}\label{sec:background}

\subsection{Algorithms for Local \textit{Post Hoc} Explanation}\label{sec:posthoc-background}
Whereas \textit{ante hoc} explainers have an intrinsic notion of interpretability, \textit{post hoc} methods serve as a surrogate explainer for a black box.
There are several classes of \textit{post hoc} explanation methods, including salience maps,
local surrogate models,
counterfactuals~\cite{DBLP:conf/ecai/WhiteG20},
and global interpretation techniques.
A comprehensive overview can be found in~\cite{guidottiSurveyMethodsExplaining2018,molnar2019,DBLP:journals/corr/abs-2010-10596}.
However, here we strictly focus on model agnostic local approximation, which is one of the most prevalent explanation strategies.
Local \textit{post hoc} explainers act as a surrogate to a black box model and
estimate the feature importance for a single decision.
In contrast, global explainers provide explanations of a model for an entire dataset.
With the ability to query the model, explainers aim to recover the local model response about an instance while isolating the most important features to produce comprehensible explanations. These methods may additionally use a \textit{background dataset}, which could come from the train or test datasets, to estimate statistics about the data and model.
It should be noted that local explanations can also be used in aggregate to gain a global understanding of a model.

We view the explanations provided by these explainers under the \textit{feature-additive} perspective, \textit{i.e.}, that explanations comprise a set of \textit{contributions} for each feature such that their sum approximates the model output~\cite{camburuCanTrustExplainer2019}. In contrast, under the \textit{feature-selectivity} perspective, explanations provide the minimal subset of features from an instance such that the model output virtually does not change (thus identifying the relevant features).
Specifically, we consider the \LIME{}~\cite{ribeiroWhyShouldTrust2016}, \SHAP{}~\cite{lundbergUnifiedApproachInterpreting2017}, \SHAPR{}~\cite{aasExplainingIndividualPredictions2020}, \PDP{}~\cite{Friedman2001}, and \MAPLE{}~\cite{plumbModelAgnosticSupervised2018} explainers.
Previously, \LIME{} has been shown to produce unstable explanations for similar inputs~\cite{shankaranarayanaALIMEAutoencoderBased2019,elshawiInterpretabilityMachineLearningbased2019}, and an analysis indicated that it can produce inaccurate explanations~\cite{pmlr-v108-garreau20a}.
In an adversarial context, \LIME{} and \SHAP{} have been shown to be deceivable by employing an out-of-distribution detector to mask prejudiced decisions~\cite{slackFoolingLIMESHAP2020}.
Moreover, a recent study has demonstrated that \textit{post hoc} explanations often disagree with one another and that practitioners have no principled approach to resolving such disagreement~\cite{krishna2022disagreement}.

\noindent
\paragraph{\underline{P}artial \underline{D}ependence \underline{P}lots}
\PDP{}s~\cite{Friedman2001} estimate the average marginal effect of a subset of features on the output of a model using the Monte Carlo method. When the subset comprises one or two features, the model output is plotted as a function of the feature values. \PDP{}{}s give a global understanding of a model, but can also yield a local explanation for the specific feature values of a sample.

\noindent
\paragraph{\underline{L}ocal \underline{I}nterpretable \underline{M}odel-agnostic \underline{E}xplanations}
\LIME{}~\cite{ribeiroWhyShouldTrust2016} explains by learning a linear model from a randomly sampled neighborhood around z-score normalized instances. Feature selection is controlled by hyperparameters that limit the total number of features used in approximation, such as the top-$k$ largest-magnitude coefficients from a ridge regression model.

\noindent
\paragraph{\underline{M}odel \underline{A}gnostic Su\underline{p}ervised \underline{L}ocal \underline{E}xplanations}
\MAPLE{}~\cite{plumbModelAgnosticSupervised2018} employs a tree ensemble, \textit{e.g.}, a random forest, to estimate the importance (the net impurity) of each feature. Feature selection is performed upfront on the background data by iteratively adding important features to a linear model until error is minimized on held out validation data. For local explanations, \MAPLE{} learns a ridge regression model on the background data distribution with samples weighed by the tree leafs relevant to the explained instance.

\noindent
\paragraph{\underline{Sh}apley \underline{A}dditive Ex\underline{p}lanations}
\SHAP{}~\cite{lundbergUnifiedApproachInterpreting2017} takes a similar but distinct approach from \LIME{} by approximating the Shapley values of the conditional expectation of a model.
Feature selection is controlled using a regularization term.
Note that when we write \SHAP{}, we are specifically referring to Kernel \SHAP{}, which is distinguished from its other variants for trees,
structured data, etc.
An extension of \SHAP{} to handle dependent features has also been proposed~\cite{aasExplainingIndividualPredictions2020}, which we refer to as \SHAPR{} after the associated \texttt{R} package. In effort to improve the accuracy of \SHAP{} explanations, \SHAPR{} estimates the conditional distribution assuming features are statistically dependent.
\subsection{Evaluation of Local \textit{Post Hoc} Explanation Algorithms}\label{sec:posthoc-eval-background}
\paragraph{General Metrics}
Some works gauge effectiveness by human evaluation and proxy tasks, including subjective ratings and the improvement of human performance~\cite{fengWhatCanAI2019,tintarevEvaluatingEffectivenessExplanations2012}.
Several classes of hand-crafted metrics have been put forward.
Those based on erasure
attempt to evaluate feature importance by removing parts of an input based on the explanation, \textit{e.g.}, under the assumption that higher-weighted features affect accuracy more~\cite{deyoungERASER2020,bachPixelWiseExplanationsNonLinear2015}.
``Fidelity'' metrics
compare surrogate accuracy relative to the model being explained~\cite{kennyTwinSystemsExplainArtificial2019,guidotti2018local}.
Another assumption made is that explanations should be robust, where better explanations are considered to be those that do not fluctuate between similar instances~\cite{alvarez-melisRobustnessInterpretabilityMethods2018}.
Further, some metrics reward monotonic behavior between feature importance and the model response~\cite{nguyenQuantitativeAspectsModel2020}.
More extensive surveys of evaluation methods covering all classes of explainers, \textit{e.g.}, counterfactual and visualization approaches, are presented in~\cite{zhouEvaluatingQualityMachine2021,Vermeire2022,pawelczyk2021carla,DBLP:journals/adac/RamonMPE20}.
While these metrics are suitable for real-world applications and many classes of explainers, they are all built upon heuristics rather than a mechanistic and principled interpretation of explanations. For instance, an accurate explainer does not imply faithfulness, a plausible explanation may only be persuasive, and unexpected explanations may arise due to the decision-making process of the model rather than the explainer. On the contrary, our work derives the actual feature-additive ground truth explanations for several classes of models in order to evaluate explainers objectively.

\paragraph{Ground Truth Evaluation}
More relevant to our study, several works have proposed the use of ground truth explanations to evaluate explainers.
The approach is far more interrelated with the model than the aforementioned metrics and a significant advance in explainer evaluation.
Guidotti
developed a framework for the evaluation of \textit{post hoc} explanations of synthetic models using ground truth~\cite{guidottiEvaluatingLocalExplanation2021}.
For tabular data, synthetic ground truth explanations are defined as the gradient with respect to each feature. They are then compared with the local explanations by an explainer.
However, this definition can often be misleading. Consider the case where the derivative with respect to a feature is negative, but the contribution that feature has to a decision is positive; this opposes the expectation of an explainer that produces explanations in terms of feature contributions.
Furthermore, gradient-based attributions have been shown to be susceptible to intrinsic noise~\cite{anconaGradientBasedAttributionMethods2019}.
In our work, we derive the true additive contributions of features
and allow for experiments on real-world problems.

In \cite{zhouFeatureAttributionMethods2021}, a dataset modification procedure is proposed to coerce a model to rely on certain features. These modified features are used as the ground truth with justification that they must be more important to the model, as long as they increase accuracy.
While this notion of ground truth is well-defined, it deviates from ours as it is derived by proxy means of feature importance.
Similarly, in~\cite{camburuCanTrustExplainer2019}, an explanation verification framework is introduced for evaluating local \textit{post hoc} explainers under the {feature-selectivity} perspective.
The framework uses the RCNN rationale generator~\cite{DBLP:conf/emnlp/LeiBJ16} to produce ground truth explanations; the subset of features in the rationale is simply the explanation. They evaluate several \textit{post hoc} explainers on NLP tasks. Our work instead approaches evaluation from the feature-additive point of view and is compatible with broader modalities of data.

\section{Framework for Evaluating \textit{Post Hoc} Explainers}\label{sec:method}

To realize proxy-less explainer evaluation, we construct a framework that has explainers attempt to explain a broad class of white box models.
These models have an unambiguous notion of ground truth that has direct equivalence relations to all considered explainers.
We also introduce a novel algorithm to enable the maximally fair comparison of ground truth and explanations.
Figure~\ref{fig:frameworka} shows a high-level overview of the framework.

\subsection{White Box \& Explainer Formulation}\label{sec:model-formulation}
\begin{figure}%
    \centering
    \begin{subfigure}[t]{0.48\linewidth}
        \centering
        \includegraphics[width=.97\linewidth]{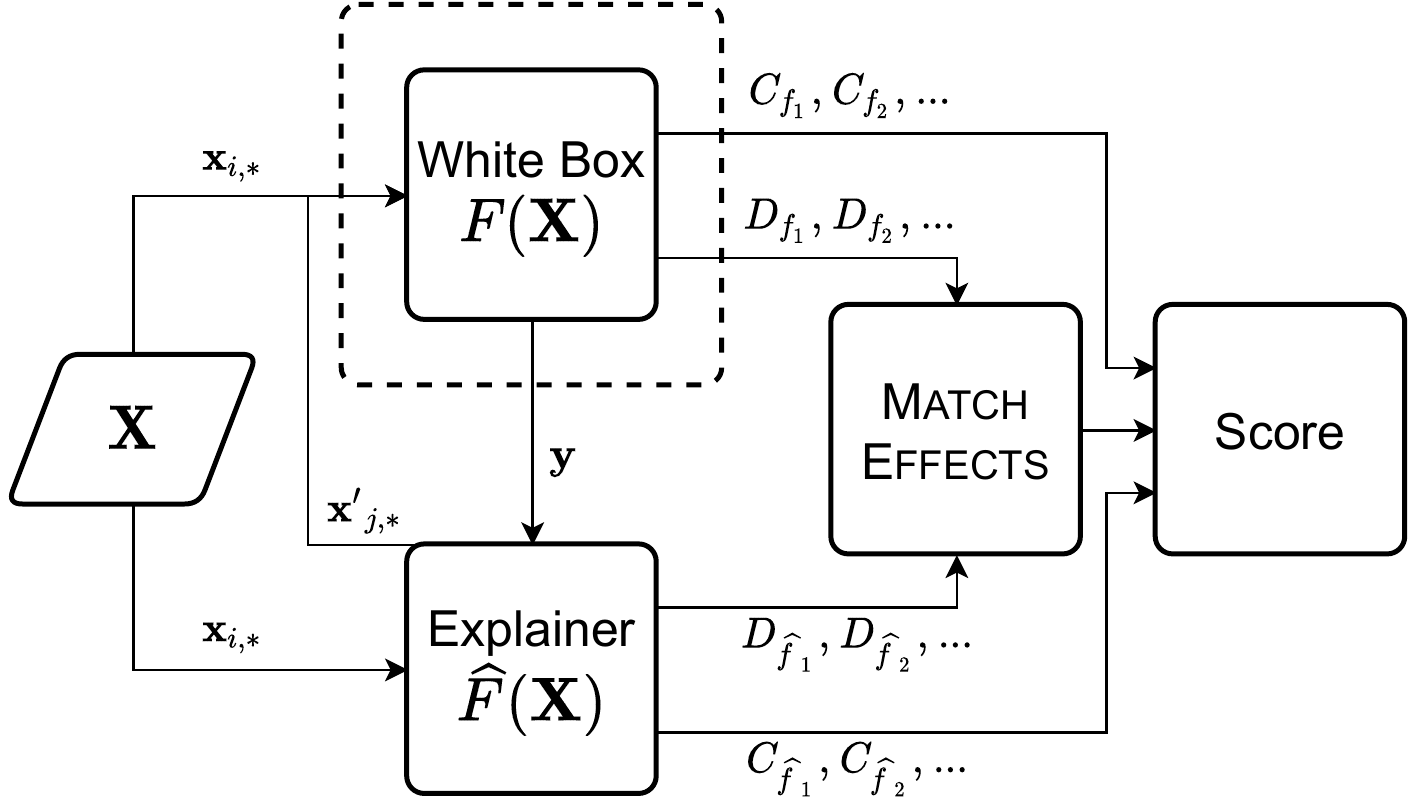}
        \caption{} \label{fig:frameworka}
    \end{subfigure}%
    \hfill
    \begin{subfigure}[t]{0.48\linewidth}
        \centering
        \includegraphics[width=.69\linewidth]{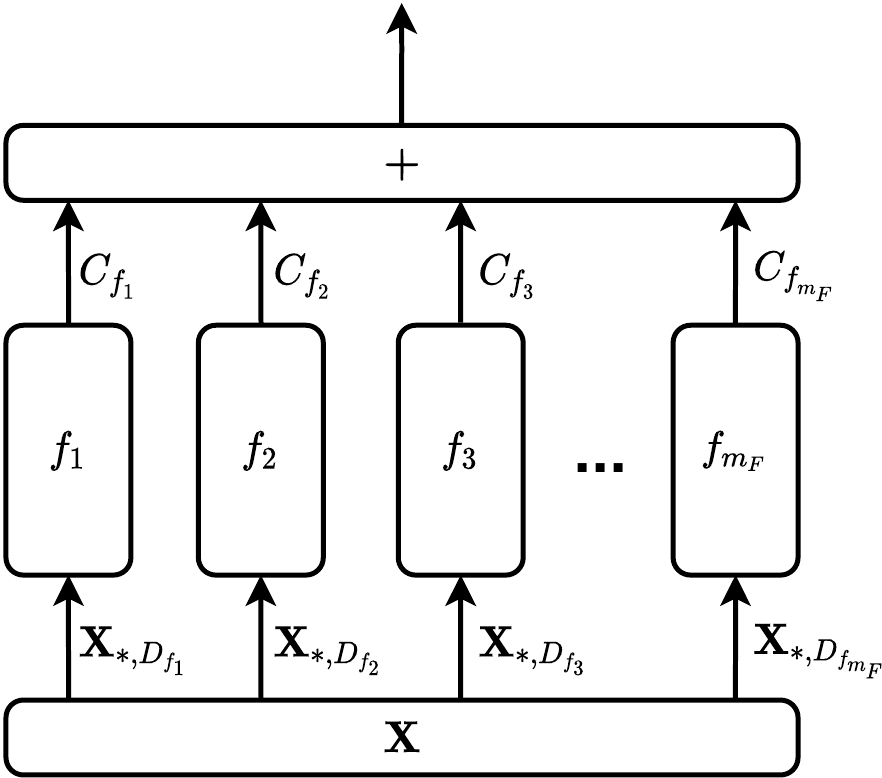}
        \caption{} \label{fig:whitebox}
    \end{subfigure}%
    \vspace{-2ex}
    \caption{(a) High-level overview of the evaluation framework. An explainer estimates the feature contributions of a white box model. These are then compared with the true feature contributions of the model with aid from the \textsc{MatchEffects} algorithm (see Section~\ref{lab:gt_align}). (b) White box model. The output is modeled as the sum of arbitrarily complex functions that transform the features within each effect.}
    \label{fig:framework}
\end{figure}

We consider a general form of white box models, similar to, but distinct from, generalized additive models (GAMs).
The additive structure
leads to concise definitions of feature contributions
while still allowing for
feature interactions, high dimensionality, and highly nonlinear effects. Note that, here, white box does not imply human-comprehensibility, but rather that
precise feature contributions can be extracted automatically.
Let $\mathbf{X} \in \mathbb{R}^{n \times d}$ be a matrix with $n$ samples and $d$ features, $D = \{i \mid 1 \le i \le d \}$ be the set of all feature indices, and $F(\cdot)$ be an additive function comprising $m_F$ effects. Each effect $f_j(\cdot)$ is a non-additive function that takes a subset of features $D_{f_j} \subseteq D$ as input and yields an additive contribution $C_{f_j}$ to the model output, as shown in Equation~\eqref{eq:additive_model}.
\begin{equation}\label{eq:additive_model}
    F(\mathbf{X}) = \sum_j^{m_F} f_j\left(\mathbf{X}_{*,D_{f_j}}\right) = \sum_j^{m_F} C_{f_j}
\end{equation}
The ground truth explanation is then the set of effects and their contributions $\{(D_{f_j}, C_{f_j}) \mid 1 \le j \le m_F\}$.
Figure~\ref{fig:whitebox} illustrates this formulation, which notably differs from GAMs in that each $f_j(\cdot)$ can be non-smooth.

For explainers, we denote local estimates of the model as $\hat{F}(\cdot)$ which comprise a summation of $m_{\hat{F}}$ effects $\hat{f}_k(\cdot)$. Similarly, the explanation from an explainer has the form $\{(D_{\hat{f}_j}, C_{\hat{f}_j}) \mid 1 \le j \le m_{\hat{F}}\}$. The explainers evaluated in this work have $m_{\hat{F}} \le d$, however, explainers with $m_{\hat{F}} > d$ are compatible with both the framework formulation and implementation.

\paragraph{Synthetic Models}
We generate synthetic models with controlled degrees of sparsity, order of interaction, nonlinearity, and size. This parameterization allows us to study how different model characteristics affect explanation quality. Here, each $f_j(\cdot)$ is a composition of random non-additive unary and/or binary operators for a random subset of features $D_{f_j}$.
Expressions are generated based on these parameters and we verify that the domains and ranges are in $\mathbb{R}$.
For example, a generated expression with $m_F = d = 4$ and 2 dummy features could look like $F(\mathbf{X}) = \mathbf{x}_{*,1} + e^{\mathbf{x}_{*,4}} +
\log(\mathbf{x}_{*,1} \mathbf{x}_{*,4})+ \frac{\mathbf{x}_{*,4}}{\mathbf{x}_{*,1}}$.
See
Appendix D
for details of our algorithm used to generate such models.

\paragraph{Learned Models}
We consider two types of learned models: GAMs and sparse neural networks.
The former was introduced as a rich but simple model: capable of modeling nonlinear effects while conducive for understanding feature significance~\cite{DBLP:books/lib/HastieTF09}. Each $f_j(\cdot)$ is a smooth nonparametric function that is fit using splines. A link function relates the summation of each $f_j(\cdot)$ to the target response, such as the identity link for regression and the logit link for classification.

The sparse neural networks we consider have the same additive structure, but each $f_j(\cdot)$ is instead a fully-connected neural network (NN). Each NN can have any architecture, operates on $D_{f_j}$, and yields a scalar value for regression or a vector for classification. The output is the summation of each NN with a link function similar to the GAM.
This structure is related to the neural additive model proposed in~\cite{agarwalNeuralAdditiveModels2020}.
This NN formulation also holds for convolutional NNs~(CNNs), which can have a non-unary $m_F$ as long as the receptive field at any layer does not cover the full image.
Notably, while the CNN operates on the image data $\bm{\mathcal{X}}\in\mathbb{R}^{n\times d_1\times d_2 \times d_3}$, the explainers operate on the flattened data $\mathbf{X}\in\mathbb{R}^{n\times d_1 d_2 d_3}$.
See Appendix A for more details.

The number of effects $m_F$ and each effect $D_{f_j}$ are selected randomly for learned models such that $m_F > 1$, the number of matches from \textsc{MatchEffects} (introduced in Section~\ref{lab:gt_align}) is $> 1$, and the task error is satisfactorily low.

\subsection{Ground Truth Alignment: \textsc{MatchEffects}}\label{lab:gt_align}

\begin{figure}
    \begin{center}
      \begin{subfigure}{0.95\linewidth}
        \includegraphics[width=\linewidth]{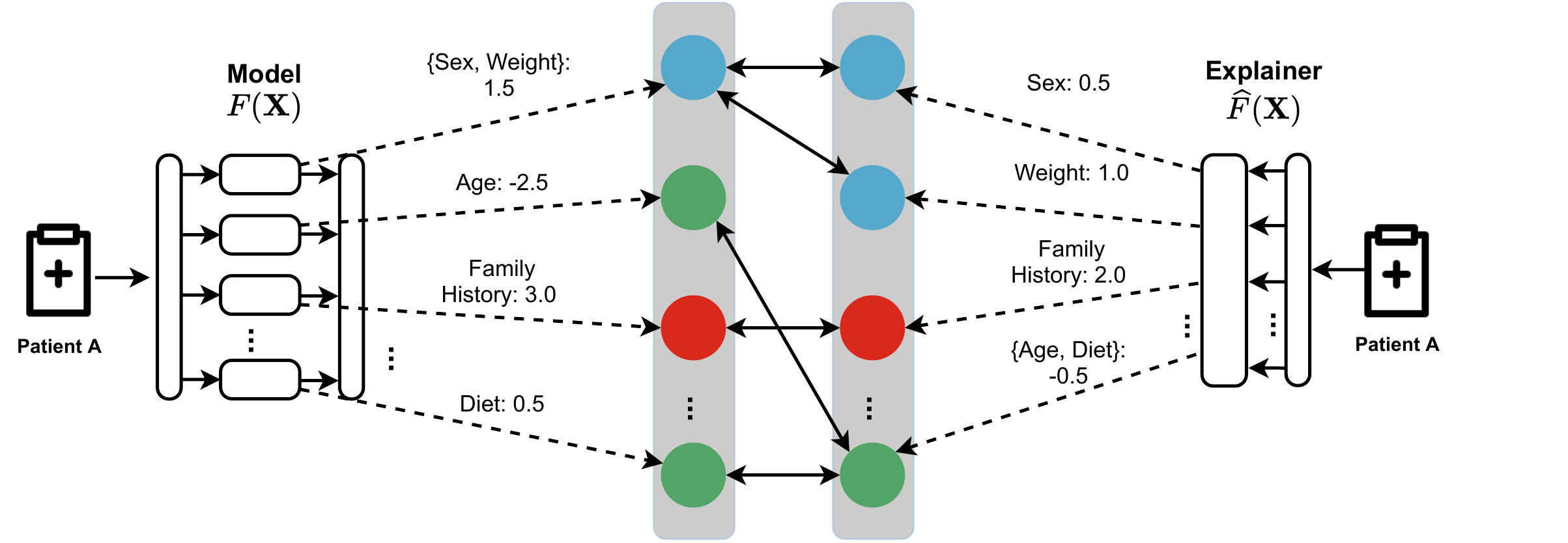}
        \caption{} \label{fig:matcheffectsex}
      \end{subfigure}\\%
      \begin{subfigure}{0.23\linewidth}
        \includegraphics[width=\linewidth]{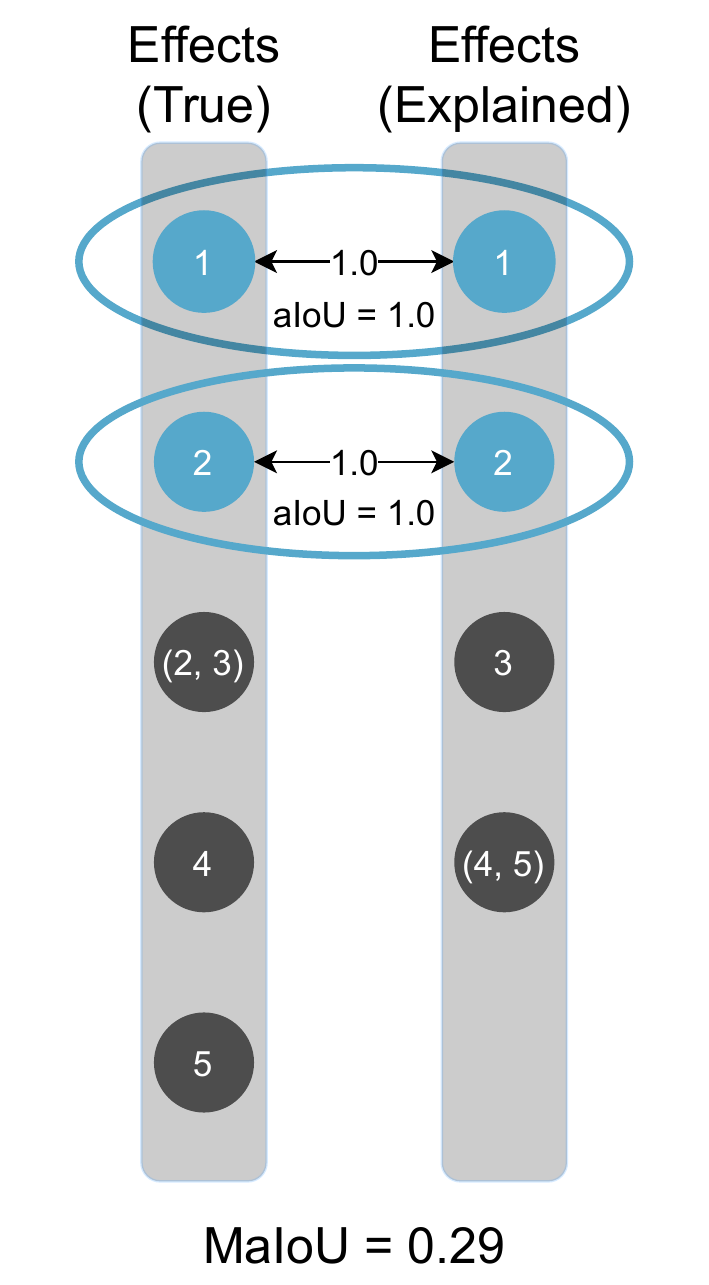}
        \caption{} \label{fig:matcheffectsa}
      \end{subfigure}%
      \begin{subfigure}{0.23\linewidth}
        \includegraphics[width=\linewidth]{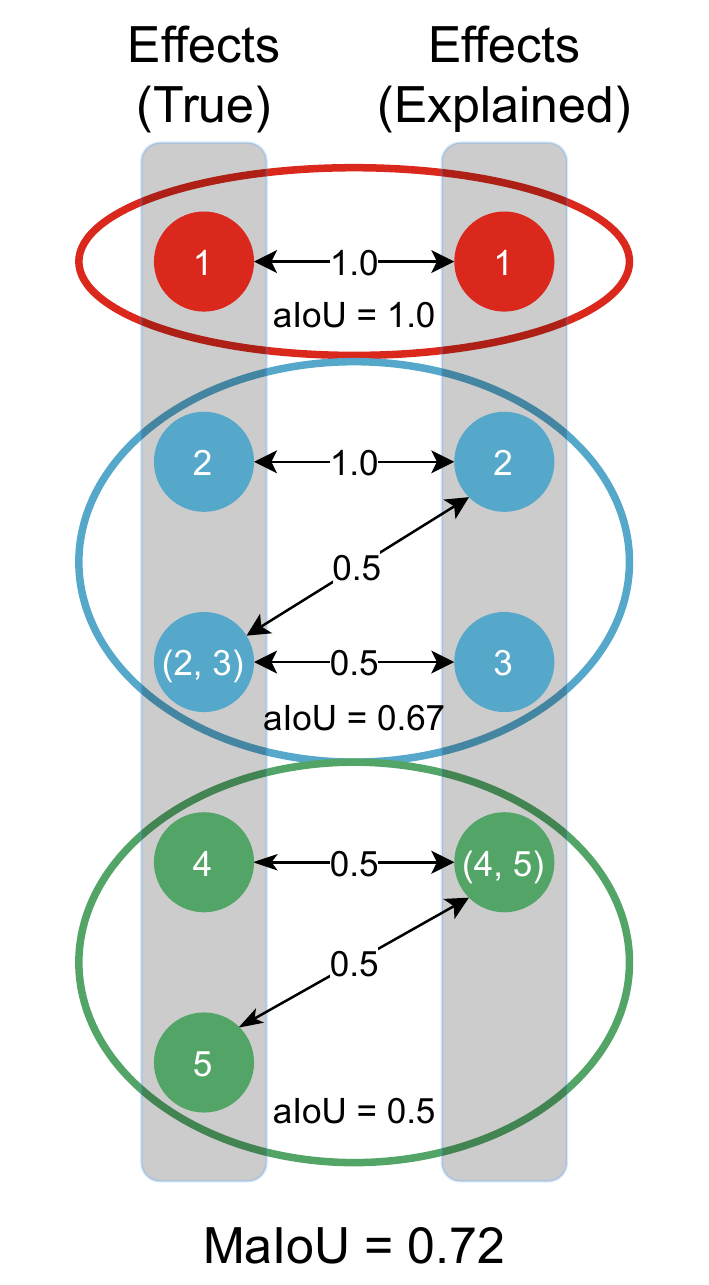}
        \caption{} \label{fig:matcheffectsb}
      \end{subfigure}%
      \begin{subfigure}{0.23125\linewidth}
        \includegraphics[width=\linewidth]{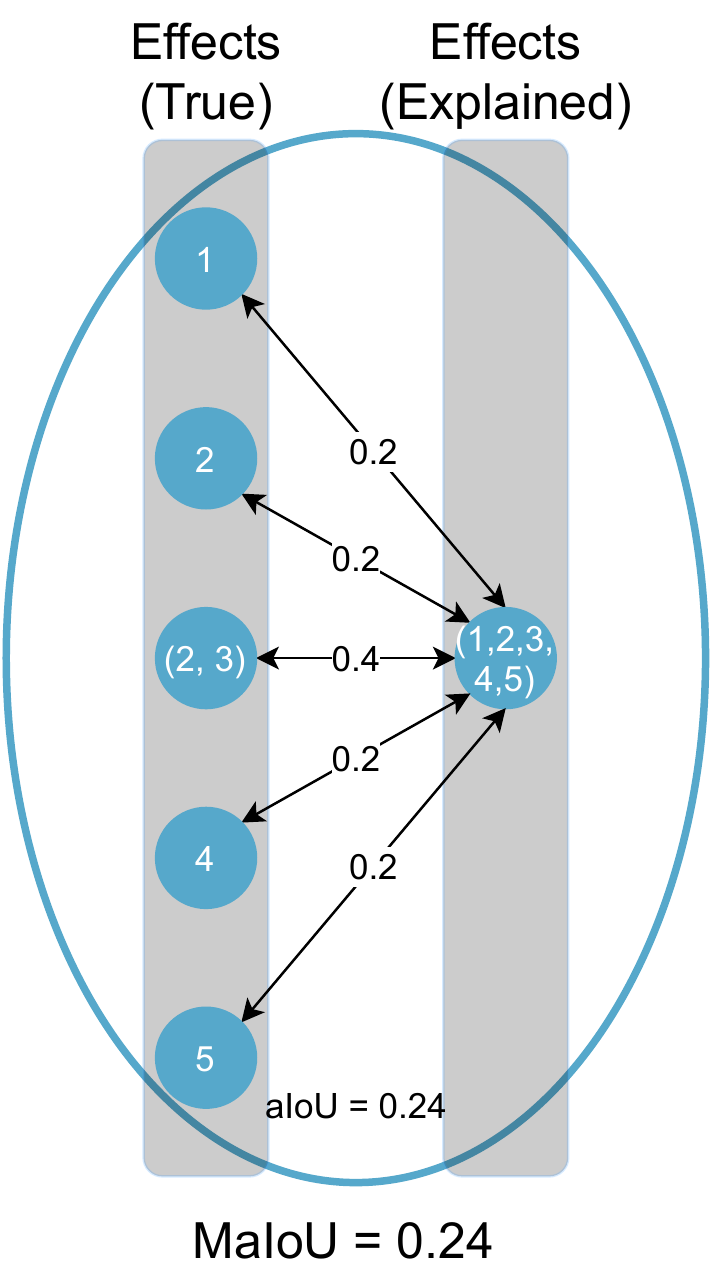}
        \caption{} \label{fig:matcheffectsc}
      \end{subfigure}
   \end{center}
   \vspace{-2ex}
    \caption{Examples of \textsc{MatchEffects} and MaIoU (Equation~\ref{eq:maiou}) in facilitating fair comparison between two sets of explanations. (a) A simple example demonstrating \textsc{MatchEffects} for a hypothetical medical task. Like colors from each side of the graph can be directly compared. While no explainer evaluated in this work explicitly detects interactions ($|m_{\hat{F}}| \ge 2$), the visual demonstrates how such explainers are compatible with the framework.
    (b) A strict matching between effects severs partially correct effects from comparison and yields a harsh MaIoU. (c) \textsc{MatchEffects} fairly groups together effects for comparison and gives a more reasonable MaIoU -- ideally, the sum of the true contributions is equivalent to the sum of the explained contributions in each group (component). (d) Importantly, MaIoU defines the goodness of a match, in this case indicating that a superficially perfect explanation is uninformative and incorrect.}
    \label{fig:matcheffects}
    \vspace{-2ex}
\end{figure}

With the framework formalism, we now have a model and an explainer, each of which produces explanations as a set of effects and their corresponding contributions.
Because there may not be a one-to-one correspondence between the two sets,
we cannot directly compare the effects.
Consider the case of a model with an interaction effect, \textit{i.e.}, some $|D_{f_j}| \ge 2$; if the explanation has no $D_{\hat{f}_k} = D_{f_j}$, then a direct comparison of explanations is not possible.
To this end, we propose the \textsc{MatchEffects} algorithm, 
which matches subsets of effects
between the model and explainer.
The main motivation for doing so is to allow models to have non-additive effects (which is the case for most model classes).

To achieve this matching, we consider all $D_{f_j}$ and $D_{\hat{f}_k}$ to be the left- and right-hand vertices, respectively, of an undirected bipartite graph.
Edges are added between effects with common features; we consider effects to be atomic and respect the non-additive structure where it exists in the model being explained. Thus, contributions may not be given for individual features.
We then find the connected components of this graph to identify groups of effects with inter-effect dependencies. 
If every component contains an exact match, for example, if $match_F = \{\{2\}, \{2, 3\}\}$ and $match_{\hat{F}} = \{\{2\}, \{2, 3\}\}$, then each contribution by $\{2\}$ and $\{2, 3\}$ will be compared separately.
\textsc{MatchEffects} is formalized in Algorithm~\ref{alg:paireffects} and illustrated for a few examples in Figure~\ref{fig:matcheffects}.
This process guarantees
the most fair and direct comparison
of explanations, and does not rely on
gradients,
sensitivity,
or other proxy means~\cite{guidottiEvaluatingLocalExplanation2021,deyoungERASER2020,fengWhatCanAI2019}.
The worst-case time complexity of \textsc{MatchEffects} is
$\mathcal{O}\left(m_F m_{\hat{F}}d\right)$
and the
space complexity is
$\mathcal{O}\left(m_F m_{\hat{F}}\right)$
(see
Appendix B
for proofs).
It should be noted that in all practical use cases, the wall-time and memory bottlenecks of the framework arise from the explainers, especially those that scale combinatorially with $d$.

\begin{algorithm}%
    \small
    \KwIn{$D_F = \{D_{f_j} \mid 1 \le j \le m_{F}\}$, the set of feature subsets operated on by
    model $F$}
    \KwIn{$D_{\hat{F}} = \{D_{\hat{f}_k} \mid 1 \le k \le m_{\hat{F}}\}$, the set of feature subsets operated on by
    explainer $\hat{F}$}
    \KwResult{Corresponding sets of effects that can be compared}
    \tcp{add edges between effects with mutual features}
    $E \gets \text{new array}$\;
    \For{$D_{f_j} \in D_F$}{
        \For{$D_{\hat{f}_k} \in D_{\hat{F}}$}{
            \If{$\lvert D_{f_j} \cap D_{\hat{f}_k} \rvert > 0$}{
                $E$.append($\{D_{f_j}, D_{\hat{f}_k}\}$)\;
            }
        }
    }
    $V \gets D_F \cup D_{\hat{F}}$\tcp*[r]{effects are vertices}
    $G \gets (V, E)$\;
    \tcp{find connected components ($CCs$) for the undirected graph $G$}
    $CCs \gets \textsc{ConnectedComponents}(G)$\;
    $matches \gets \text{new array}$\;
    \tcp{$V_c$ and $E_c$ comprise the vertices and edges of component $c$, respectively}
    \For(\tcp*[f]{unpack the components}){$\{V_c, E_c\} \in CCs$}{
        $match_F \gets \text{new array}$\;
        $match_{\hat{F}} \gets \text{new array}$\;
        \For{$D_c \in V_c$}{
            \eIf{$D_c \in D_{F}$}{
                $match_F$.append($D_c$)\;
            }{
                $match_{\hat{F}}$.append($D_c$)\;
            }
        }
        \If{$match_F = match_{\hat{F}}$}{
            \tcp{elements of identical sets are each a perfect match}
            \For{$D_c \in match_F$}{
                $matches$.append($\{\{D_c\}, \{D_c\}\}$)\;
            }
        }{
            $matches$.append($\{match_F, match_{\hat{F}}\}$)\;
        }
        
    }
    \KwRet{$matches$}
    \caption{\textsc{MatchEffects}}
    \label{alg:paireffects}
\end{algorithm}

One could exploit \textsc{MatchEffects} by producing explanations that attribute the entire output of the model to a single effect comprising all $d$ features; the comparison of contributions could trivially yield perfect but uninformative scores. Likewise, a model with such interaction effects, like most deep NNs, would render this evaluation inconsequential.
To mitigate this issue, we introduce a metric that evaluates the goodness of the matching.
Let $E_c$ be the set of edges of a single component found by \textsc{MatchEffects}. For an edge $\{D_{f_j}, D_{\hat{f}_k}\} \in E_c$, the intersection-over-union (IoU), also known as the Jaccard index, is calculated between $D_{f_j}$ and $D_{\hat{f}_k}$. The total goodness for a component is the average of the IoU scores of each edge in $E_c$, and the total goodness for a matching is the mean value of these averages: the mean-average-IoU (MaIoU). This metric is given by Equation~\eqref{eq:maiou}
\begin{align}\label{eq:maiou}
    \text{MaIoU}(CCs) &= \frac{1}{|CCs|} \sum_{\{V_c, E_c\} \in CCs}
    \Bigg(
        \frac{1}{|E_c|} \sum_{\{D_{f_j}, D_{\hat{f}_k}\} \in E_c} \frac{|D_{f_j}\cap D_{\hat{f}_k}|}{|D_{f_j} \cup D_{\hat{f}_k}|}
    \Bigg)%
\end{align}
where $CCs$ is defined in Algorithm~\ref{alg:paireffects}.
MaIoU can be thought of the degree to which the effects uncovered by an explainer agree with the true effects of the model. Figure~\ref{fig:matcheffectsb} shows the effectiveness of MaIoU on an example with three components, and Figure~\ref{fig:matcheffectsc} shows how MaIoU can inform when an explanation is uninformative and mechanistically incorrect, mitigating the aforementioned consequences.

\subsection{Equivalence Relations to Explainers}\label{sec:equiv-relation}

With \textsc{MatchEffects} and MaIoU defined, a direct comparison between true and explained explanations is nearly possible.
However, some adaptation is still required due to the use of normalization and differing definitions of ``contribution'' between explainers. Here, we bridge together these definitions.
\LIME{} normalizes the data as z-scores, \textit{i.e.}, $z = (x_i - \mu_i) / \sigma_i$, before learning a linear model. We then need to scale the coefficients $\Theta = \{\theta_i \mid 1 \le i \le d\}$ of each local linear model using the estimated means $\mu_i$ and standard deviations $\sigma_i$ from the data as follows.
\begin{multicols}{2}%
  \noindent\begin{align}%
\theta_0' &= \theta_0 - \sum_i \frac{\mu_i \theta_i}{\sigma_i}%
\end{align}%
  \begin{align}%
\theta_i' &= \frac{\theta_i}{\sigma_i}%
\end{align}%
\end{multicols}
\noindent
In \SHAP{}, the notion of feature importance is the approximation of the mean-centered independent feature contributions for an instance. The expected value $\mathbb{E}[F(\mathbf{X})]$ is estimated from the background data \SHAP{} receives. In order to allow for valid comparison, we add back the expected value of the true contribution $\mathbb{E}[C_{f_i}]$ estimated from the same data. However, since a 1:1 matching is not a guarantee, we must consider all effects grouped by said matching:
\begin{align}
    C_{match_{\hat{F}}} &= \sum_{k \in match_{\hat{F}}} \hat{f}_k(\mathbf{x}_{*,k}) + \sum_{j \in match_{F}} \mathbb{E}[C_{f_j}]
\end{align}
The same procedure applies to \SHAPR{}.
See
Appendix B
for the derivations of these relations.

Furthermore, \LIME{} and \MAPLE{} provide feature-wise explanations as the coefficients $\Theta$ of a linear regression model. In turn, we must simply compute the product between each coefficient and feature vector $\mathbf{x}_{*,i} \,\theta_i$ to yield the contribution to the output according to the explainer.

\section{Experimental Results}\label{sec:results}

We evaluate the explainers on thousands of synthetic problems and popular real-world datasets.
By varying the data and models, we identify when explainers fail, whether plausible explanations are faithful, and other interesting trends.
\paragraph{Setup}

We use
{\small\texttt{SymPy}}~\cite{sympy}
to generate synthetic models and represent expressions symbolically as expression trees. This allows us to automatically discover the additivity of arbitrary expressions.
See
Appendix A
for the unary and binary operators, parameters, and operation weights considered in random model generation.
All stochasticity is seeded for reproducibility, and all code is documented and open-sourced%
\footnote{The source code for this work is available at
\href{https://github.com/craymichael/PostHocExplainerEvaluation}{github.com/craymichael/PostHocExplainerEvaluation}%
}.
The framework is implemented in 
{\small\texttt{Python}}~\cite{van1995python}
with the help of {\small\texttt{SymPy}} and the additional libraries
{\small\texttt{NumPy}}~\cite{2020NumPy-Array},
{\small\texttt{SciPy}}~\cite{2020SciPy-NMeth},
{\small\texttt{pandas}}~\cite{pandas},
{\small\texttt{Scikit-learn}}~\cite{sklearn},
{\small\texttt{Joblib}}~\cite{joblib},
{\small\texttt{mpmath}}~\cite{mpmath},
{\small\texttt{pyGAM}}~\cite{pygam},
{\small\texttt{PDPbox}}~\cite{PDPbox},
{\small\texttt{alibi}}~\cite{alibi},
{\small\texttt{TensorFlow}}~\cite{tensorflow2015-whitepaper},
{\small\texttt{Matplotlib}}~\cite{matplotlib},
and
{\small\texttt{seaborn}}~\cite{seaborn}.
Furthermore, we build a {\small\texttt{Python}} interface to the
{\small\texttt{R}}~\cite{R}
package
{\small\texttt{shapr}}~\cite{shapr}
using
{\small\texttt{rpy2}}~\cite{rpy2}.
Appendix A
also details hyperparameters used to train the GAMs and sparse NNs, and the hardware used to run experiments.
Last, we consider the explanation of an effect to be $0$ if every estimated contribution is within a tolerance\footnote{See the documentation of \href{https://numpy.org/doc/1.19/reference/generated/numpy.allclose.html}{\texttt{numpy.allclose}} for details}. This is a fairer evaluation and tends to favor the explainers in experiments when dummy variables are present.

\subsection{Evaluation}\label{sec:evaluation}
We measure the error between the explanations of the ground truth and explainer using a few metrics.
The set of contributions comprising each explanation can be thought of as a vector collectively, thus we can compute the distance between them after applying \textsc{MatchEffects}. First considered is Euclidean distance to understand the magnitude of the disagreement with the ground truth. To quantify the disagreement in orientation (rotation), we utilize cosine distance (\textit{i.e.}, $1 - \text{cosine similarity}$), which is bounded by the interval $[0,2]$ with our data -- a distance of one indicates orthogonal contribution vectors and a distance of two indicates contribution vectors pointing in opposite directions. We also consider normalized (interquartile) root-mean-square error (\textsc{nrmse}) for comparing individual effects, as defined by Equation~\eqref{eq:nrmse}
\begin{equation}\label{eq:nrmse}
    \text{\textsc{nrmse}}(\mathbf{a}, \mathbf{b}) = \frac{1}{Q_3^\mathbf{a} - Q_1^\mathbf{a}} \sqrt{\frac{\sum_i^n (a_i - b_i) ^ 2}{n}}
\end{equation}
\noindent
where $Q_3^\mathbf{a}$ and $Q_1^\mathbf{a}$ are the third and first quartiles of $\mathbf{a}$, respectively.

\begin{figure*}%
    \centering
    \begin{subfigure}{\linewidth}
        \includegraphics[width=\linewidth]{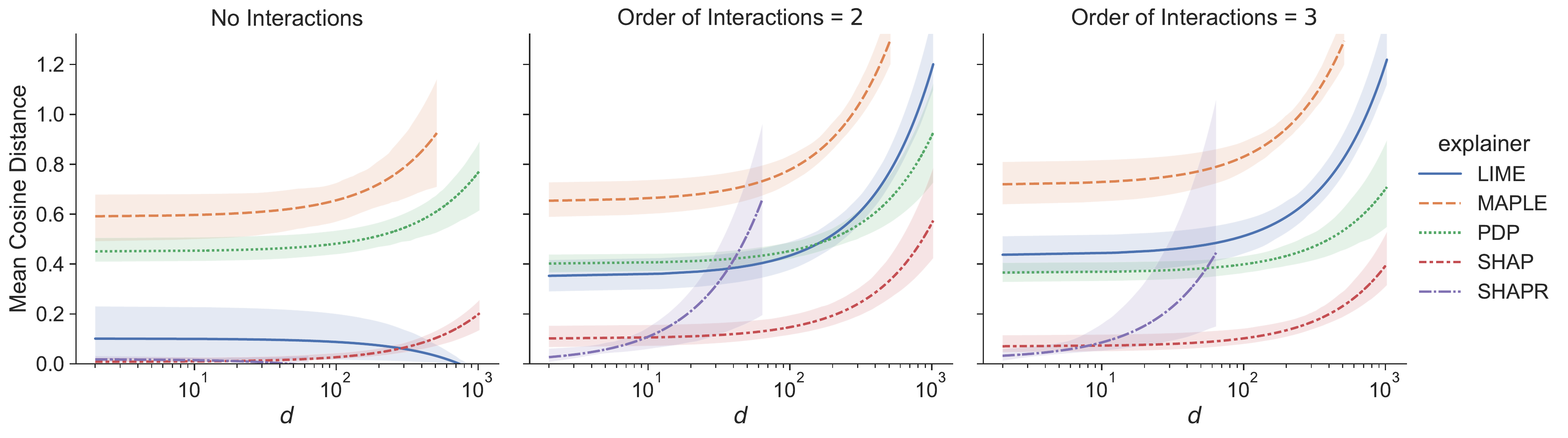}
        \label{fig:synthetic_1a}
      \end{subfigure}\\%
      \vspace{-3ex}
      \begin{subfigure}{\linewidth}
        \includegraphics[width=\linewidth]{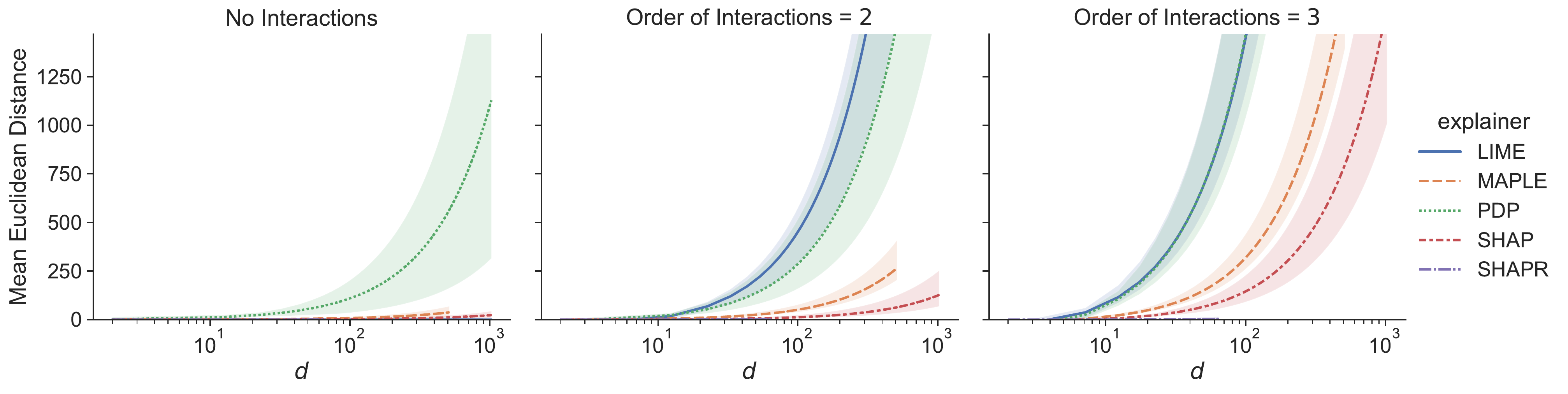}
        \label{fig:synthetic_1b}
      \end{subfigure}%
    \vspace{-6ex}
    \caption{Average cosine distances (top) and Euclidean distances (bottom) between ground truth and explained effect contributions as a function of $d$, the number of features in the synthetic dataset. A robust linear regression fit is plotted on top of the scatter plot with a 99\% confidence interval estimated with 1,000 bootstrap resamples. The y-axis has an upper limit of the 99\% percentile for visualization purposes.%
    }
    \label{fig:synthetic_1}
\end{figure*}

\subsection{Synthetic Problems}\label{sec:synth-problems}
We first demonstrate our framework on 2,000 synthetic models that are generated with a varied number of effects, order of interaction, number of features, degree of nonlinearity, and number of unused (dummy) variables. For each, we discard models with invalid ranges and domains that do not intersect with the interval $[-1, 1]$. The data of each feature $\mathbf{x}_{*,i}$ is sampled independently from a uniform distribution $\mathcal{U}(-1, 1)$. We draw $n$ samples quadratically proportional to the number of features $d$ as $n = 500 \sqrt{d}$.
The explainers are evaluated on each model with access to the full dataset and black box access to the model.

Of the 2,000 models, 16 were discarded due to the input domain producing non-real numbers. Furthermore, the explainers were not able to explain every model due to invalid perturbations and resource exhaustion\footnote{See Appendix A for hardware and time budgets.}.
The former occurred with \PDP{}, \LIME{}, and \SHAP{}, typically due to models with
narrower feature domains, while the latter occurred with \MAPLE{} and \SHAPR{} due to
the inefficient use of background data and intrinsic computational complexity.
In total, 82\%, 39\%, 80\%, 91\%, and 40\% of models were successfully explained by \PDP{}, \LIME{}, \MAPLE{}, \SHAP{}, and \SHAPR{}, respectively.
We consider the failure to produce an explanation for valid input to be a limitation of an explainer or its implementation.
Results demonstrate the efficacy of the proposed framework in understanding explanation quality, as well as factors that influence it when paired with the experimental design.
As the dimensionality, the degree of interactions, and the number of interactions increase, the disagreement between ground truth and explanation grows.
Figure~\ref{fig:synthetic_1} illustrates these results for all of the explained synthetic models.
Because \LIME{} failed to explain a substantial portion of synthetic models, it appears to improve with an increased $d$ in the leftmost plots;
in reality, it only succeeded in explaining simpler models with a larger $d$.
\SHAP{} performs the best relative to the other explainers, maintaining both a closer and more correctly-oriented explanation compared to the ground truth.
Interestingly, the ranking of \LIME{} and \MAPLE{} swaps when comparing average cosine and Euclidean distances.
Whether more similar orientation or magnitude is desired depends upon the application.
Appendix C
includes additional experiments with synthetic models, including evaluation as a function of the number of number of interactions, number of nonlinearities, and dummy features.
\subsection{Real-World Case Studies}\label{sec:real-problems}

\begin{table}%
    \small
    \centering
    \ra{1.1}
    \begin{tabular}{@{}lccccccccc@{}}
        \toprule
        \multirow{2}{*}{Dataset} & \multirow{2}{*}{Model} && \multicolumn{5}{c}{Explainer Error} && \multirow{2}{*}{$\rho_{perf}$}\\
        \cmidrule{4-8}
        &&& {\PDP{}} & {\LIME{}} & {\MAPLE{}} & {\SHAP{}} & {\SHAPR{}} \\
        \midrule
        \multirow{2}{*}{Boston}
            & GAM && 0.340 & 0.709 & 0.652 & 0.001 & 0.111 && 0.995 \\
            &  NN && 0.278 & 0.182 & 0.431 & 0.001 & 0.209 && 0.351 \\[.9ex]
        \multirow{2}{*}{COMPAS}
            & GAM && 0.821 & 0.781 & 0.863 & 0.000 & -- && 0.800 \\
            &  NN && 0.328 & 0.062 & 0.274 & 0.001 & -- && 1.000 \\[.9ex]
        \multirow{2}{*}{FICO}
            & GAM && 0.795 & 0.949 & 0.962 & 0.003 & -- && 0.200 \\
            &  NN && 0.761 & 0.193 & 0.270 & 0.001 & -- && 0.800 \\[.9ex]
        \multirow{1}{*}{MNIST}
            & CNN && 0.660 & 0.253 & 0.318 & 0.049 & 0.175 && 0.410 \\
        \bottomrule
    \end{tabular}
    \vspace{-2ex}
    \caption{Real-world explainer results on several datasets for GAMs and NNs. Here, explainer error is the cosine distance averaged over all samples and classes, if applicable. $\rho_{perf}$ is Spearman's rank correlation coefficient between the mean explanation cosine distance and explainer accuracy. \SHAPR{} is not implemented for data with categorical variables in this work.}
    \label{tab:real_world_results}
\end{table}

Incomprehensible models applied to real-world problems require transparency when the stakes are high. Local explanations on this type of data need to be faithful to the model, otherwise they can propagate spurious relationships.
To this end, we evaluate GAMs and sparse NNs on several real-world datasets.

The Boston housing dataset~\cite{HARRISON197881} contains median home values in Boston, MA, that can be predicted by several covariates, including sensitive attributes, \textit{e.g.}, those related to race. Models that discriminate based off of such features necessitate that their operation be exposed by explanations.
We also evaluate explainers on the Correctional Offender Management Profiling for Alternative Sanctions~(COMPAS) recidivism risk dataset~\cite{angwin2016machine}.
The dataset was collected by ProPublica in 2016 and contains
covariates, such as criminal history and demographics, the proprietary COMPAS risk score, and recidivism data
for defendants from Broward County, Florida.
The FICO Home Equity Line of Credit~(HELOC) dataset~\cite{ficoheloc2018}, introduced in a 2018 XAI challenge, is also used in this work. It comprises anonymous HELOC applications made by consumers requesting a credit line in the range of \$5,000 and \$150,000. Given the credit history and characteristics of an applicant, the task is to predict whether they will be able to repay their HELOC account within two years.
Last, we evaluate on a down-sampled version of the MNIST dataset~\cite{MNIST}. With the aim of reducing explainer runtime and improving comprehensibility of effect-wise results, we crop and then resize each handwritten digit in the dataset
to $12\times 10$ and only include a subset of the 10 digits. See Appendix A for more details.

\begin{figure}%
    \centering
    \begin{subfigure}{.63\linewidth}
        \includegraphics[width=\linewidth]{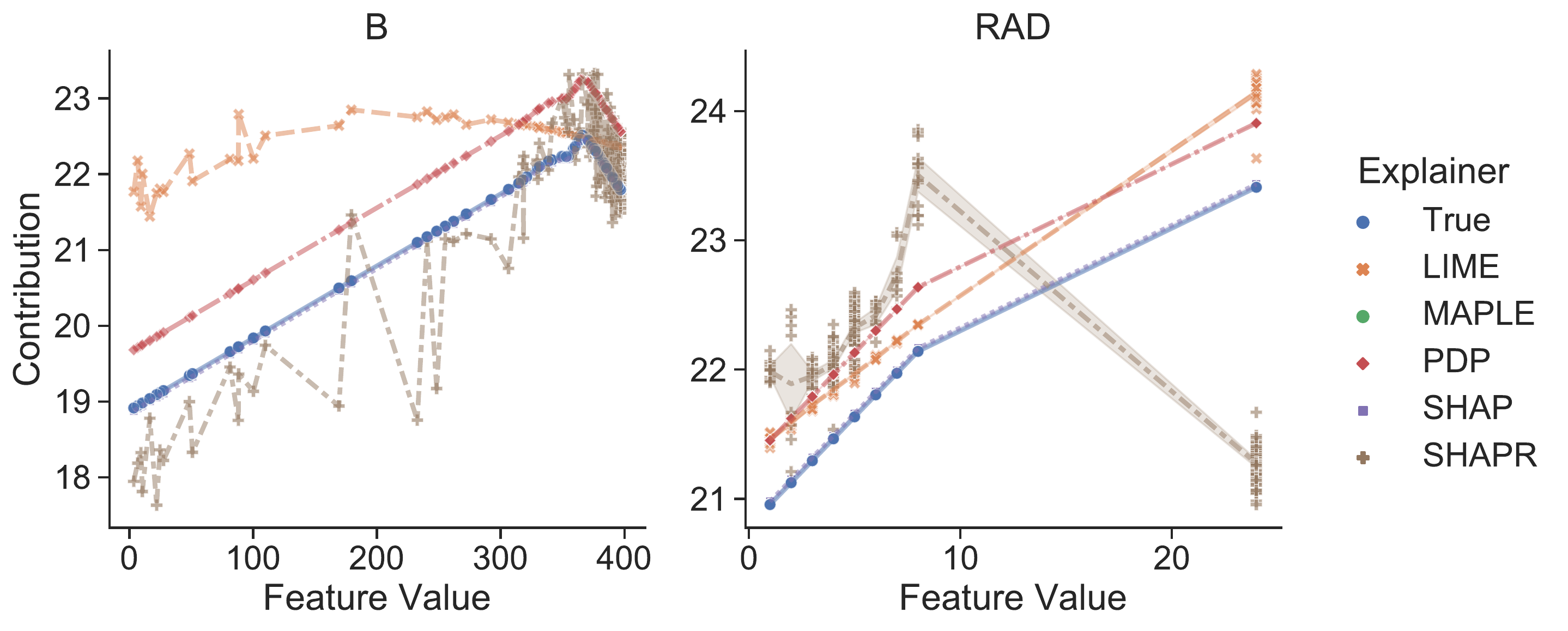}
        \caption{}
      \end{subfigure}%
      \begin{subfigure}{.355\linewidth}
        \includegraphics[width=\linewidth]{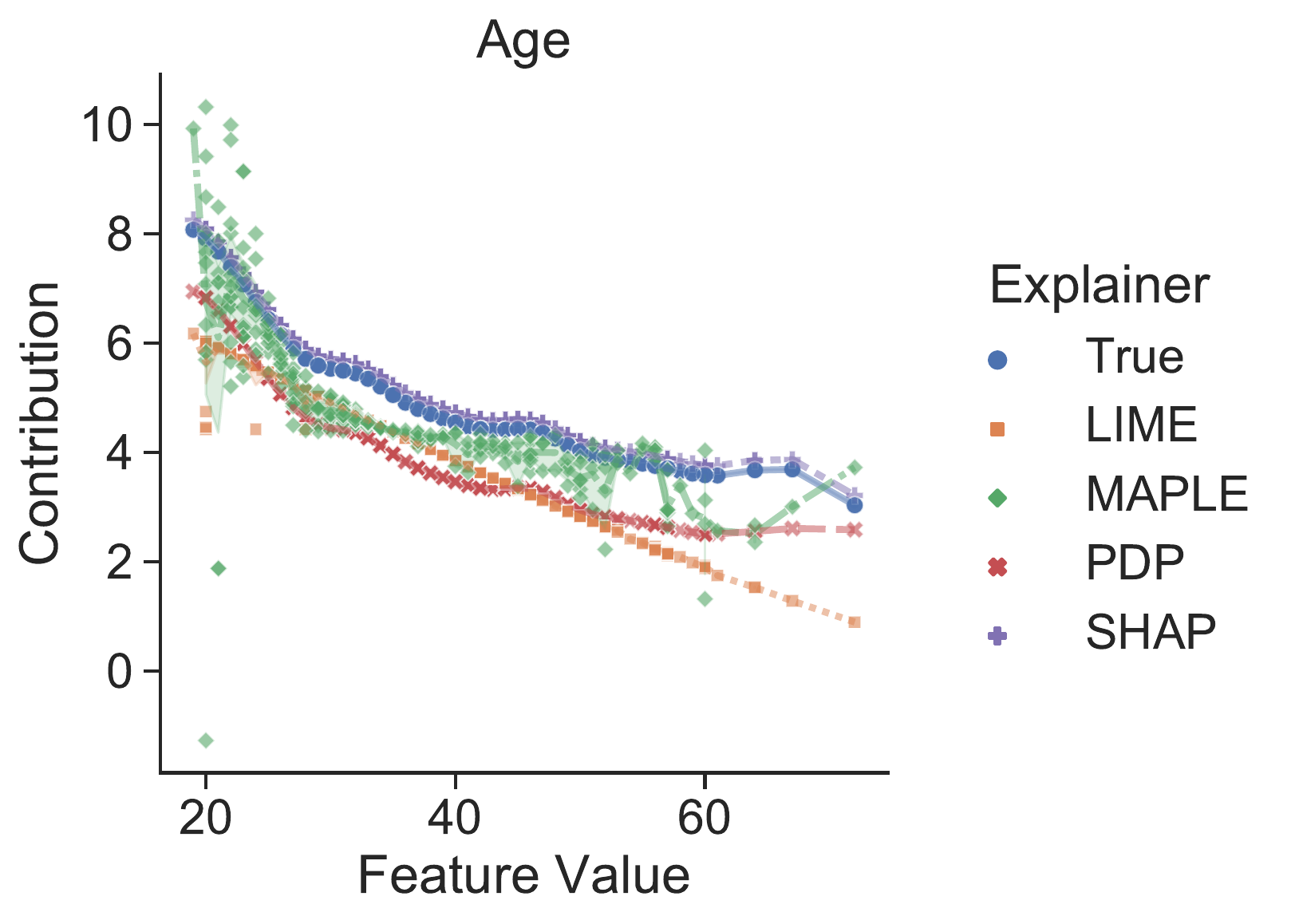}
        \caption{}
      \end{subfigure}%
    \vspace{-2ex}
    \caption{The true and explained feature shapes of (a) RAD (index of accessibility to radial highways) and B (proportion of African American population by town\protect\footnotemark) from a NN trained on the Boston housing dataset, and (b) Age from a GAM trained on the COMPAS dataset.}
    \label{fig:boston_subset}
\end{figure}

Table~\ref{tab:real_world_results} contains the aggregate results across all real-world datasets for the considered models. Among the considered explainers, \SHAP{} outperforms on all datasets and models, often by several orders of magnitude.
Surprisingly, \SHAPR{} performs worse than \SHAP{}, but still ranks well compared to the other explainers. \PDP{}, \LIME{}, and \MAPLE{} produce poor explanations in general, and all explainers struggled more with the GAMs than the considered NNs.
To test whether explainer fidelity correlates with accuracy, we compute the Spearman's rank correlation coefficient $\rho_{perf}$ between the mean explanation cosine distance and explainer accuracy. Recall that under the feature-additive perspective that the sum of the contributions from an explainer approximates the model output, which can be treated as the prediction of the explainer. The scores, shown in Table~\ref{tab:real_world_results}, demonstrate that a plausible explainer, \textit{i.e}, one that predicts accurately, does not necessarily produce faithful explanations, and vice versa.
The MaIoU scores for each explainer are present in Appendix~C. We discuss how feature selection and feature interactions influence the score, as well as how the score determines the completeness of explanations.

\footnotetext{While the Boston housing dataset is widely studied as a baseline regression problem, the data column (``B'') is notably controversial; the original paper~\cite{HARRISON197881} includes and preprocesses the data as $B = 1000(B' - 0.63)^2$ where $B'$ is the proportion of African Americans by town.}

We also visualize a subset of results in Figure~\ref{fig:boston_subset} as feature shapes.
This shows more clearly that several explainers do not faithfully explain some of the feature contributions.
For example, \SHAPR{}, \MAPLE{}, and \LIME{} fail to satisfactorily unearth how the proportion of African Americans living in an area (feature B), according to the NN, drive the housing price;
\SHAPR{} produces a high-variance estimate ($\text{\textsc{nrmse}} = 1.68$),
\MAPLE{} fails to even detect the effect ($\text{\textsc{nrmse}} = 3.59$),
and
\LIME{} only is able to approximate the mean contribution value ($\text{\textsc{nrmse}} = 3.04$).
This type of failure is incredibly misleading to any user and potentially damaging if the model is deployed.
Fortunately, in this instance, \SHAP{} reveals this relationship within reasonable error ($\text{\textsc{nrmse}} = 0.129$).
The COMPAS visualization shows another example of explanations of the ``Age'' feature of a GAM. Again, several explainers produce misleading and noisy explanations.

\begin{figure}[t]
    \centering
    \begin{subfigure}{0.45\linewidth}
        \includegraphics[width=\linewidth]{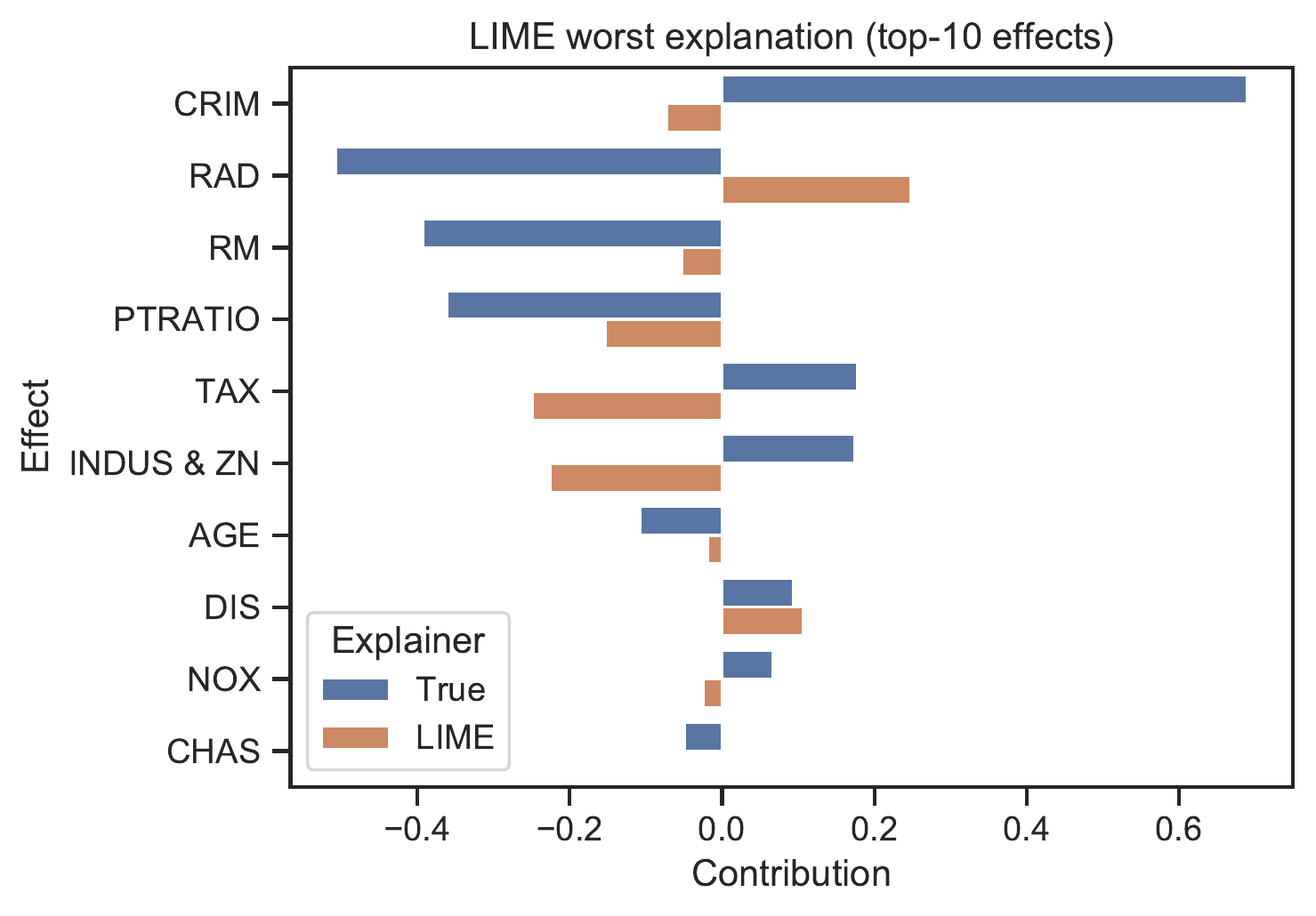}
    \end{subfigure}%
    \begin{subfigure}{0.45\linewidth}
        \includegraphics[width=\linewidth]{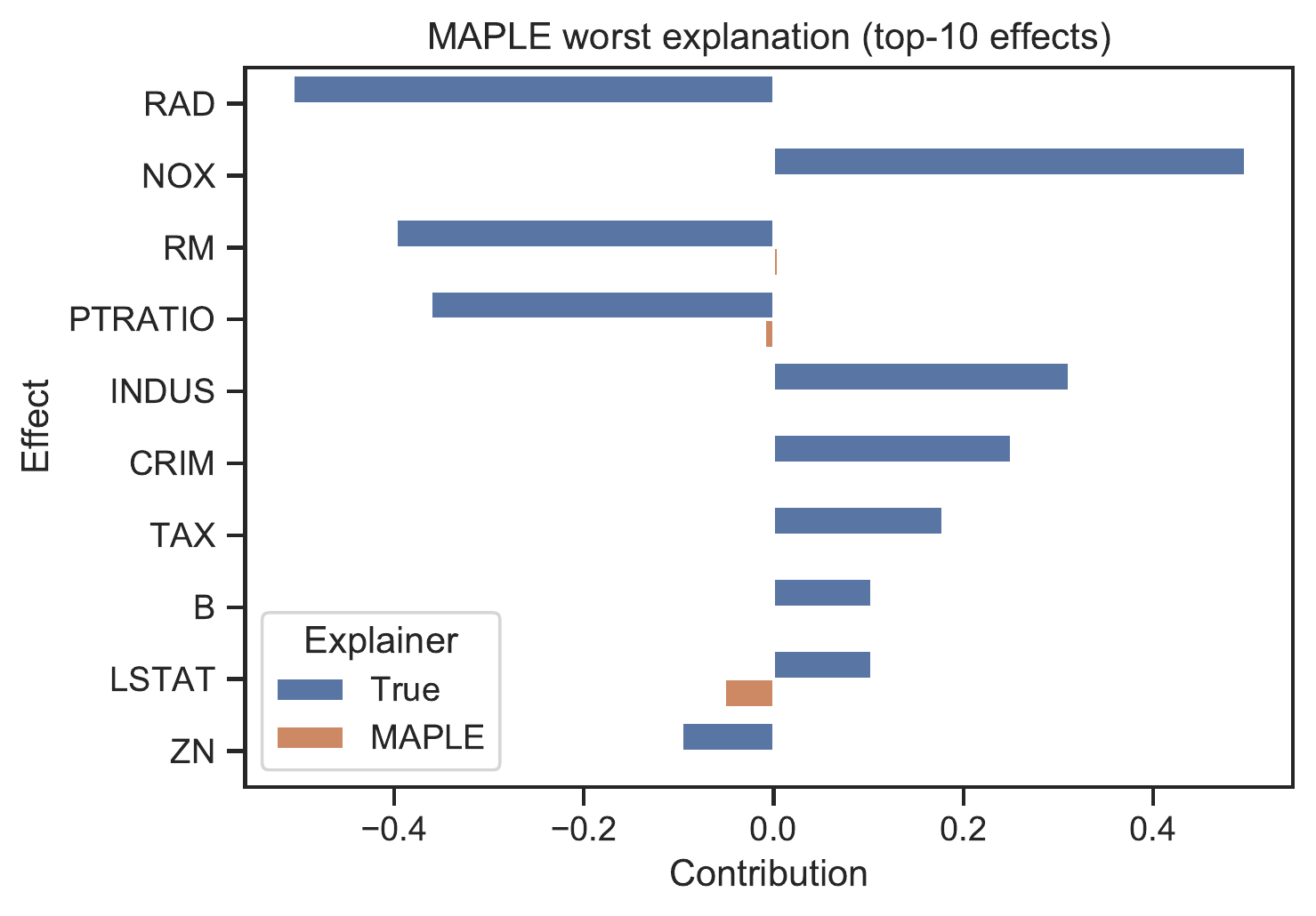}
    \end{subfigure}\\%
    \begin{subfigure}{0.45\linewidth}
        \includegraphics[width=\linewidth]{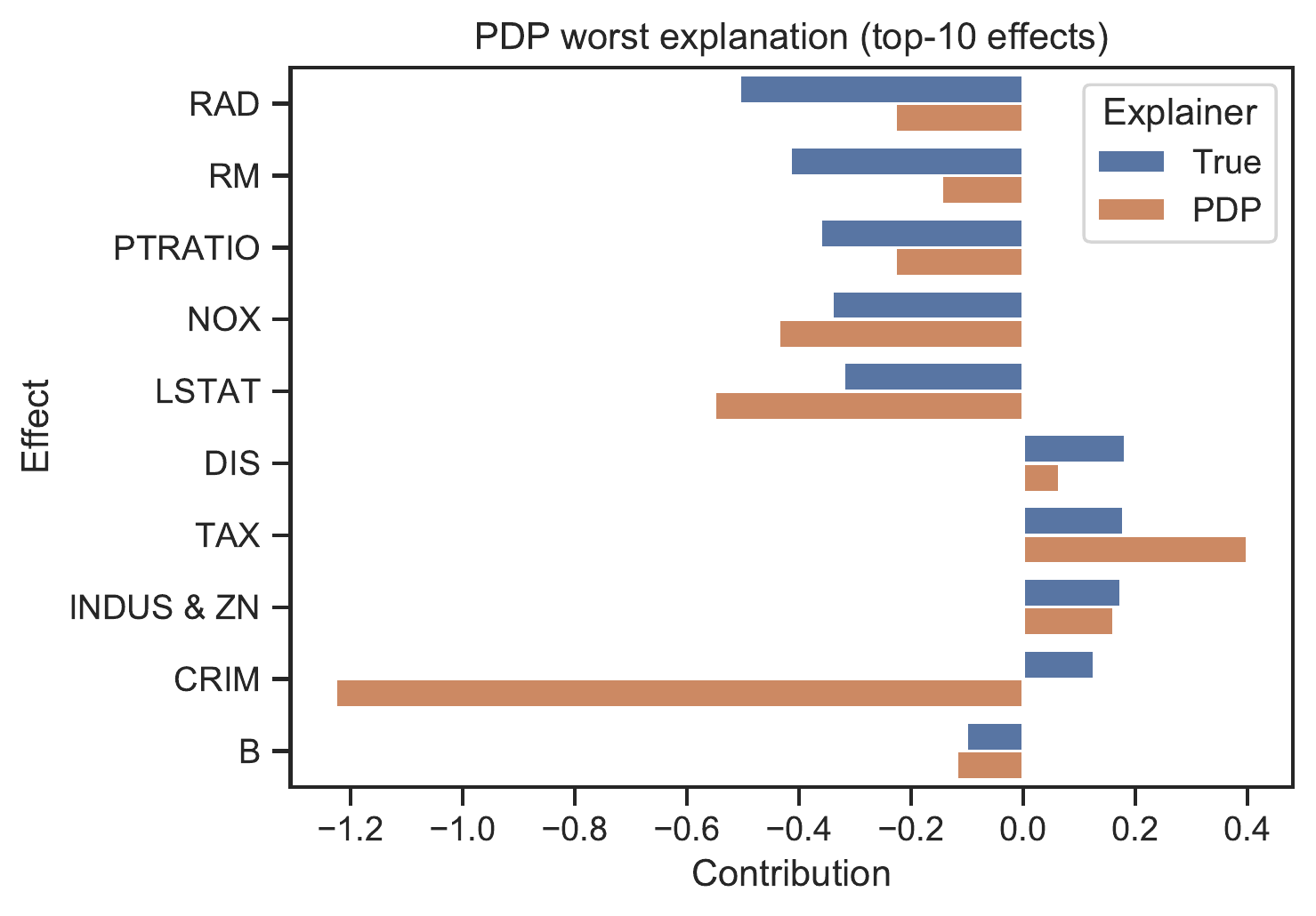}
    \end{subfigure}%
    \begin{subfigure}{0.45\linewidth}
        \includegraphics[width=\linewidth]{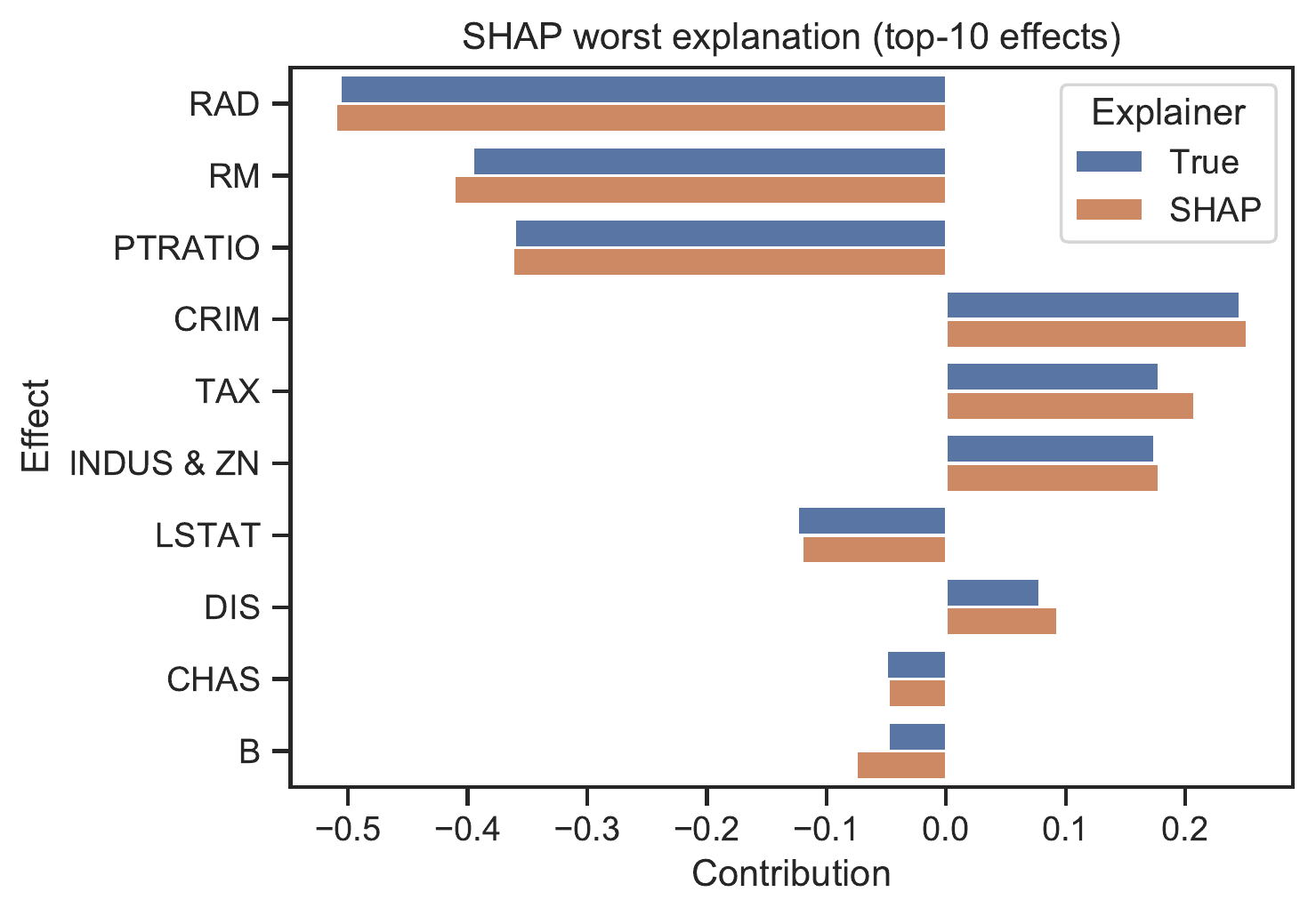}
    \end{subfigure}\\%
    \begin{subfigure}{0.45\linewidth}
        \includegraphics[width=\linewidth]{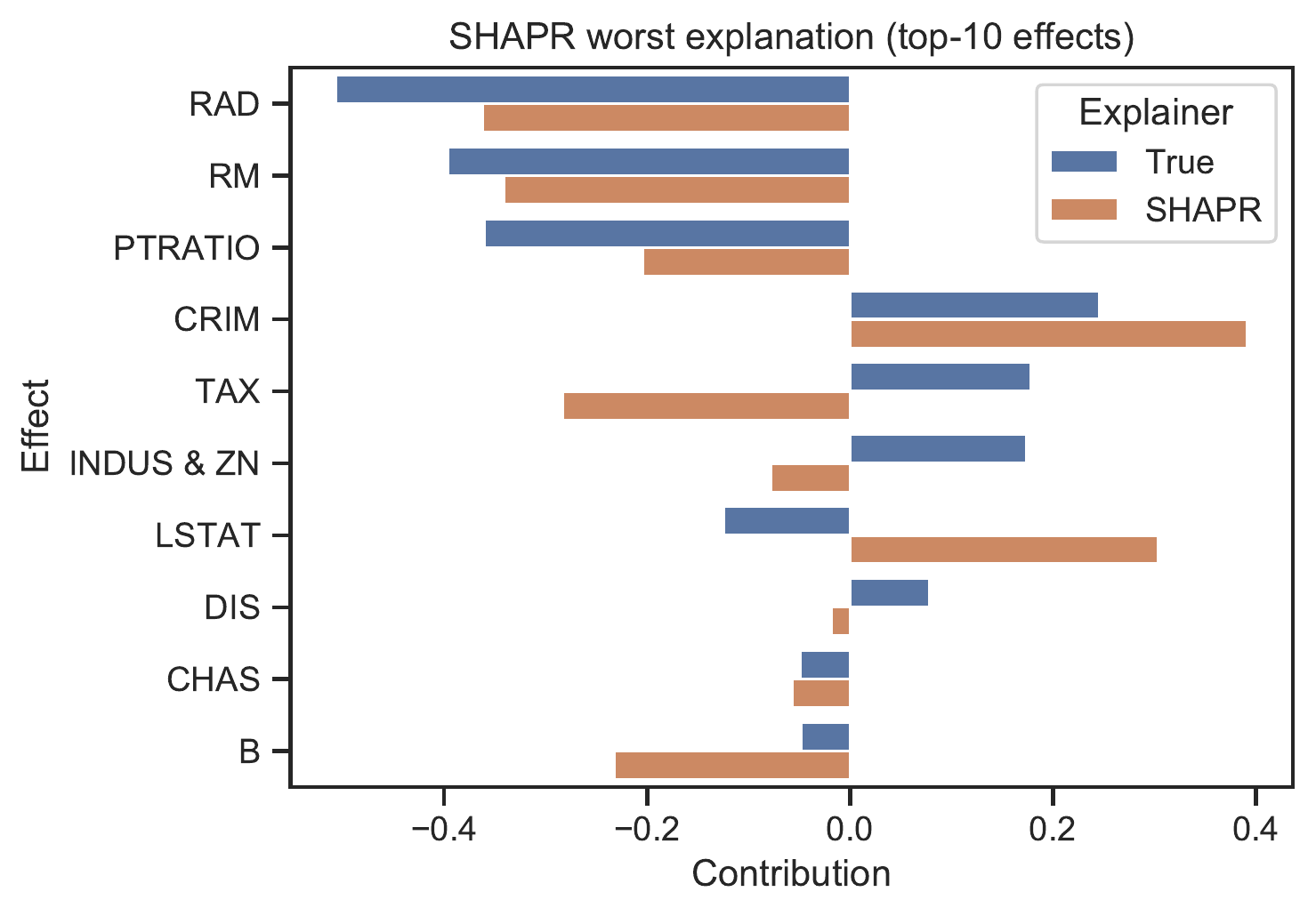}
    \end{subfigure}%
    \vspace{-2ex}
    \caption{The top-10 explained effects of the worst explanation of a GAM trained on the Boston dataset from each explainer. Top effects are ranked by magnitude and the quality of explanation is ranked by the mean cosine distance among all explained samples.}
    \label{fig:topk_worst_explanations}
\end{figure}

Looking at aggregate metrics is not sufficient to understand how poor an explanation may be. We consider the utility of an explainer in high-stakes applications to be limited by its worst explanation. Consequently, we visualize the worst explanations from each explainer and show those on the Boston dataset in Figure~\ref{fig:topk_worst_explanations}. The 10 most relevant effects to each decision are shown and the quality of the explanation is assessed using cosine distance. Again, the low point of \SHAP{} is still of relatively high quality, and the other explainers reveal incorrect attributions of effects. \MAPLE{} selects few important features correctly and does so conservatively. \PDP{}, \LIME{}, and \SHAPR{} all problematically assign opposite-signed contributions for several effects.
The remaining explanation visualizations, as well as feature shapes, heat maps, and other results, are shown in Appendix C.
\section{Discussion}\label{sec:discussion}

The proposed framework is capable of objectively evaluating explainers and discovering characteristics of models that drive infidelity.
Unlike previous work, our evaluation methodology moves beyond considering the plausibility of explanations and alleviates the need for proxy or subjective metrics.
Rather, we compare explanations to ground truth that is directly derived from the additive effects of the model being explained.
We demonstrate the capability of our framework to expose limitations of explainers; in experiments, several popular explainers produced inadequate accounts of the model decision-making process, especially when the number of features and feature interactions increased.

The shortcomings of these explainers arise from their underlying assumptions which do not hold true for the majority of models, especially the assumptions of independent features and local linearity.
These assumptions are further impacted by the explainer hyperparameters, such as the kernel width for \LIME{} or the background summarization method for \SHAP{}.
Tuning of these parameters is dependent upon the data and model. In practice, these knobs can be adjusted until the explanations ``look right,'' which is not realistic when the most faithful hyperparameters need to be derived from the black box itself.
Furthermore, while \SHAP{} performs best, the algorithm requires simplifying assumptions in order to guarantee its tractability, precluding its faithful application in many use cases~\cite{DBLP:conf/aaai/BroeckLSS21}.
This is especially troubling as studies show that data scientists overtrust or do not understand interpretability techniques~\cite{kaurInterpretingInterpretabilityUnderstanding2020,krishna2022disagreement}. With the results of our study, even those practitioners who do not abuse these explanation tools may still be mislead.

It is important to consider that a caveat of our evaluation methodology is that poor explanation methods can be identified, but it cannot prove that an explainer is faithful for all models.
Thus, this framework can be seen as a benchmark and a testing ground for explainer quality.
In addition, explainers that succeed in the evaluation will not necessarily meet the subjective apprehension by all users.
However, if an explainer proves unfaithful within our framework, then it should not be user-facing regardless of whether users trust or comprehend it.
Emphasizing some of the arguments made in~\cite{nguyenQuantitativeAspectsModel2020, zhouEvaluatingQualityMachine2021}, our framework can alleviate some of the financial and other resource burdens by filtering out explainers, \textit{e.g.}, before a user study, that do not meet satisfactory marks.
Moreover, this filtering of explainers can reduce liabilities and aid in meeting AI regulations.

\paragraph{Future Work}
A natural extension of this work would be to use the framework to evaluate additional explanation methods and guide the improvement of explainer quality.
We believe that progress within this class of explainers will emerge by accounting for interdependence between features, better defining locality, and scaling computation for high-dimensional data.
Furthermore, we intend to evaluate the quality of existing hand-crafted evaluation metrics to understand how faithful they are from our proposed feature-additive perspective.
\bibliography{bibby}

\iftrue
\clearpage
\appendix
\section*{Supplemental Material}
Due to the size of the supplemental material for this work, we have elected to provide a link instead of appending to the main paper. We highly recommend viewing the additional results and visualizations of explanations within our framework. The material is structured as follows:

\begin{itemize}
    \item \textbf{Appendix A: Reproducibility}
    \item \textbf{Appendix B: Proofs and Derivations}
    \item \textbf{Appendix C: Additional Results and Figures}
    \item \textbf{Appendix D: Synthetic Model Generation}
\end{itemize}

The material is hosted at the public URL:\\ \href{https://drive.google.com/file/d/1e4HkKPb4QRJYqnHuZu4_JFH5T1g_C0p5/view?usp=sharing}{https://drive.google.com/file/d/1e4HkKPb4QRJYqnHuZu4\_JFH5T1g\_C0p5/view?usp=sharing}\\
Alternatively, if you are reading this paper on arXiv, the material has been uploaded as an ancillary file.

\fi

\end{document}